\documentclass[]{ws-ijns} 
\usepackage{multirow}

\usepackage{xcolor}

\begin{document}



%
\title{Masked Transformer for image Anomaly Localization}

%
 \author{Axel De Nardin }
 \address{Department of Mathematics, Computer Science and Physics, Università degli Studi di Udine\\via delle Scienze 206, 33100 Udine, Italy\\
 Email: denardin.axel@spes.uniud.it}

\author{Pankaj Mishra}
 \address{Department of Mathematics, Computer Science and Physics, Università degli Studi di Udine\\via delle Scienze 206, 33100 Udine, Italy\\
 Email: mishra.pankaj@spes.uniud.it}

\author{Gian Luca Foresti}
 \address{Department of Mathematics, Computer Science and Physics, Università degli Studi di Udine\\via delle Scienze 206, 33100 Udine, Italy\\
 Email: gianluca.foresti@uniud.it}

\author{Claudio Piciarelli}
 \address{Department of Mathematics, Computer Science and Physics, Università degli Studi di Udine\\via delle Scienze 206, 33100 Udine, Italy\\
 Email: claudio.piciarelli@uniud.it}


\maketitle

\begin{abstract}
Image anomaly detection consists in detecting images or image portions that are visually different from the majority of the samples in a dataset. The task is of practical importance for various real-life applications like biomedical image analysis, visual inspection in industrial production, banking, traffic management, etc. Most of the current deep learning approaches rely on image reconstruction: the input image is projected in some latent space and then reconstructed, assuming that the network (mostly trained on normal data) will not be able to reconstruct the anomalous portions. However, this assumption does not always hold. We thus propose a new model based on the Vision Transformer architecture with patch masking: the input image is split in several patches, and each patch is reconstructed only from the surrounding data, thus ignoring the potentially anomalous information contained in the patch itself. We then show that multi-resolution patches and their collective embeddings provide a large improvement in the model's performance compared to the exclusive use of the traditional square patches. The proposed model has been tested on popular anomaly detection datasets such as MVTec and headCT and achieved good results when compared to other state-of-the-art approaches.

\end{abstract}

\keywords{
Anomaly Detection; Vision Transformer;
Image inpainting; Self-supervised learning
}

\begin{multicols}{2}
\section{Introduction}
\label{sec:intro}
The ability to identify anomalous instances in large sets of data is of great importance in many different fields of application.
One typical example is represented by the need to detect and discard faulty products in industrial production lines\cite{Kozarmenik2020,Piciarelli2019,Tout2017AutomaticVS}. The ability to do so in an automated and effective way represents a very important problem for manufacturing companies. Another area where this type of problem is of even greater importance is the one represented by the analysis of biomedical data\cite{HanMADGAN2021,manzanera2019,Pogorelov2017,ansari2019} in which the early detection of anomalies can make the difference between life and death for a patient.

While for humans dealing with this kind of problem is a rather easy task, we cannot say the same for machines. One main reason that makes it a difficult problem to address is the fact that, while it is essentially a classification problem, classical classification approaches cannot be used because of the nature of the data analyzed. When dealing with anomaly detection problems, the data is often highly unbalanced in favor of ``normal'' instances, while we have very few examples representing the ``abnormal'' ones \cite{Piciarelli2021}.
When dealing with images we also have the added problem of the high dimensionality of the data which often leads to more classical methods for anomaly detection, such as clustering techniques, to achieve poor performance.
For this reason, many of the most recent approaches for image anomaly detection adopt a deep neural network model to map the inputs into a latent feature space to which classical approaches can be applied more effectively or, in most cases, which can be used to obtain an image reconstruction which is then compared to the original one to perform a self-supervised kind of training. The idea behind these approaches is that, since the model is usually trained only on the normal images, there should be a larger difference between the reconstructed and the original image for the anomalous instances, and therefore we should be able to identify them effectively during the testing process.

Most of the models used in recent years to tackle the anomaly detection problem make use of Convolutional layers which exploit the typical characteristics of images by detecting progressively more complex features starting from the most basic ones (e.g. edges). In particular, there are two kinds of architectures, with their respective variations, which gained a lot of popularity for their effectiveness in dealing with this kind of problem, which are Autoencoders\cite{ChenAE} and Generative Adversarial Networks (GANs)\cite{ZenatiGAN2018}.
One problem with classical image reconstruction approaches though, is that they use the information extracted from the whole image to obtain the output image, which can lead the model to be able to reconstruct also the anomalies and therefore to not be able to identify them.

In this paper, we propose a new approach a novel method for image Anomaly Detection, called Masked Transformer for image Anomaly Localization (MeTAL), that adopts as its backbone the recently presented Vision Transformer (ViT) architecture\cite{dosovitskiy2021an}, which instead of leveraging the prior knowledge granted by the use of convolutional layers, is characterized by the adoption of a masked multi-head self-attention mechanism that allows the model to learn a relationship between different patches of the input images. 
In particular, with the present work we introduce two main novelties to the original architecture. the first one regards a new masking component we added to the multi head self attention module of the ViT encoder which allows us to reconstruct each patch of the image without using any information coming from the patch itself but using only the information extracted from the surrounding patches based on the importance given to each of them by the attention module.
The second idea we present regards instead the way patches are generated from the original image. In particular, instead of relying on just the square patches as in the original work presenting ViT, we introduce the idea of calculating attention between patches of different shapes, which are then combined to obtaine the final image reconstruction.
As we will show in section 4.4 both ideas resulted in an improvement in performance over the baseline model for the task at hand.

Furthermore we show that the Vision Transformer architecture is a valid option for anomaly detection problems and can be adopted effectively even in scenarios where the amount of data available is relatively small without necessarily relying on a pre-training procedure.

The rest of the paper is organized as follows. In section~\ref{sec:related} we give an overview of other recent works related to the Anomaly detection problem, and we give a brief introduction to the Vision Transformer architecture. Then in section~\ref{sec:solution}, a detailed overview of the training process is given together with a thorough description of the proposed architecture. The obtained results are outlined in section~\ref{sec:evaluation} where a more in-depth description of the adopted dataset is also provided. Finally, in section~\ref{sec:concl} we summarize our work and discuss our ideas for future work.

\section{Related work}
\label{sec:related}
Many different approaches have been proposed to tackle the problem of Unsupervised anomaly detection and segmentation on images, both involving traditional methods and, in recent years, deep neural networks which, in most cases involve the use of convolutional modules.

In this section, we will focus mainly on the methods used as a reference by the authors of the MVTec dataset\cite{Mvtec2019} which also represent our benchmark for the present work.
The first set of approaches used for Unsupervised Anomaly Detection and based on image reconstruction is represented by GAN-based models\cite{GoodfellowGAN2014}. A GAN network is typically composed of two main components, a Generator which starting from a latent representation tries to generate images as close as possible to the one present in the dataset used, and a Discriminator which receives as input both the images generated by the Generator and the original ones and tries to discriminate between the two. 
One way of performing anomaly detection through the use of GANs is to train the model only with ``good'' images so that during the testing process it should be able to recognize them but not the anomalous one and therefore it could be used to discriminate between data with and without anomalies. A different approach, which also allows performing Anomaly localization, was proposed by Schlegl et al.\ with the introduction of the AnoGAN architecture\cite{schlegl2017unsupervised}. The main idea behind this model is represented by the introduction of an additional component which is trained to learn the inverse transformation of the Generator in order to produce the latent representation of the original images, which can then be used to perform a reconstruction of said images and thus compare it with the respective original ones.
The main issue with this idea is that the inversion process is very computationally expensive. For this reason, another model, known as f-AnoGAN\cite{SchleglFAnoGAN2019} (faster AnoGAN) was introduced. This approach, compared to the original AnoGAN network, provides much faster convergence thanks to the introduction of an additional encoder network used to learn a function that maps the original images into their respective latent space, which makes the expensive process of finding the inverse of the generator superfluous.

The second set of models, which also represents the bulk of the current approaches for unsupervised anomaly detection in images, is represented by Convolutional Autoencoders\cite{Goodfellow-et-al-2016} (AE). The idea behind the AE architecture is to use an encoder to map the inputs to a latent space that is much smaller than the original one and which is then used as the input for a Decoder which tries to learn how to reconstruct the original input starting from this latent representation. The assumption is that the model should learn to map only the most important, or more common, features regarding the instances of the dataset leaving out every superfluous or specific information. Therefore it shouldn't be able to properly reconstruct anomalies, leading to a greater difference between input and reconstruction for this class of instances compared to the normal ones. For this reason, one possible approach to anomaly detection is to use an Autoencoder, usually trained only on the normal instances, to obtain a latent representation (which is typically much smaller than the original one) for each element of the dataset and then apply a clustering algorithm in order to discriminate between the good and the anomalous ones. The limit of this approach is that it cannot be used for the localization of the anomalies but only for the detection.
Many approaches based on AEs have been proposed over time with different characteristics related to their structure and type of loss functions adopted. Recent works specifically focused on Anomaly Segmentation\cite{BergmanSSIM2019} in images, showed the benefit of using a structural similarity-based loss (e.g. SSIM) to assess the quality of the reconstruction in substitution, or addition, to the pixel-wise one (e.g. MSE) adopted in previous works.

Hereafter we present some additional approaches to anomaly segmentation developed in recent years. Napoletano et al.\cite{Napoletano2018} proposed a region-based CNN architecture, which we will refer to as CNN Feature Dictionary for consistency with the 2021 MVTec paper which we used as a reference point for this work, which extracts the training features from random patches that are cropped out of the original images and then uses a K-Means classifier to model their distribution. Since this approach provides only a binary decision on whether an image contains a defect or not, the classification process must be repeated multiple times over different patches of the image in order to obtain a spatial anomaly map, which becomes a very computationally expensive process for large images.
\begin{figure*}[htb]
    \centering
    \includegraphics[width=\linewidth]{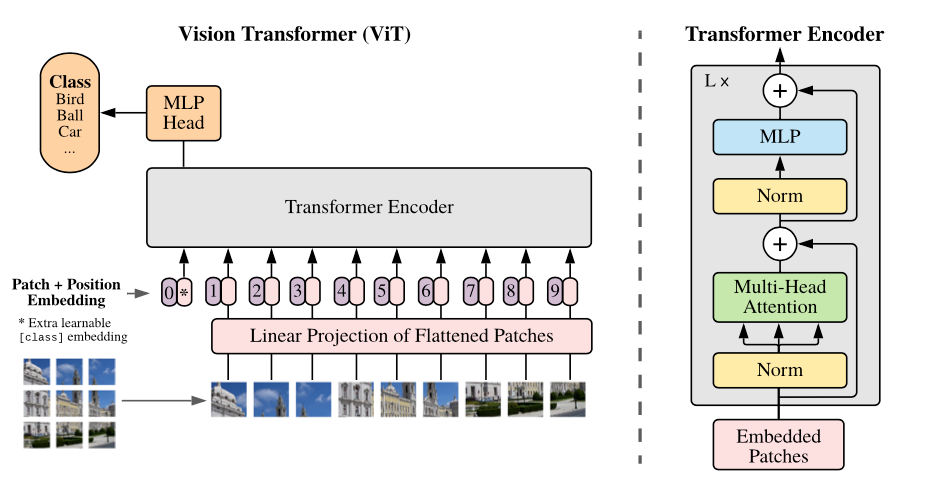}
    \caption[\cite{dosovitskiy2021an} ]{Vision Transformer architecture\footnotemark}
    \label{fig:vit}
\end{figure*}
A different approach\cite{Bergman2020}, utilizing an ensemble of randomly initialized student networks which are trained against regression targets obtained from a Teacher network pre-trained on anomaly-free data (specifically ImageNet), has been proposed under the name of \textit{student-teacher} network. The regions containing a defect are identified thanks to the fact that the student networks are not able to correctly predict the teacher's descriptor for them, thus yielding larger regression errors as well as a higher predictive uncertainty.
\footnotetext{Image taken from \url{https://github.com/google-research/vision\_transformer}}

Another interesting approach has been presented with the PIADE architecture\cite{piade2020} which focuses on multi scale feature representation through the adoption of a pretrained ResNet-18 followed by a set of pyramidal pooling layers which allow the image feature to be analyzed at different magnifications.

Latent Space Autoregression (LSA) \cite{Abati2019LatentSA} represents yet another approach to AEs for anomaly detection, combining the reconstruction loss obtained from a deep convolutional Autoencoder to the maximization of the likelihood of the latent space rapresentation through an autoregressive density estimator in order to minimize the entropy of these representations while preserving the quality of the reconstructed images. 

Variational AEs have also been adopted in works regarding the anomaly segmentation problem such as the work by Baur et al.\refcite{Baur2018} which focused on the detection of anomalous regions in Brain MR Images, where they achieved state-of-the-art performance even if not improving by a large margin over regular CAEs.

Furthermore, there are also examples of frameworks trying to combine both AEs and GANs to exploit the potential of both architecture. One such example is represented by OCGAN \cite{Perera2019OCGANON} which combines 4 main components in order to attempt to force the model to reconstruct abnormal inputs as normal ones.The 4 components are namely a denoising AE, a Latent Discriminator that forces each instance of the latent space to represent an image from the corresponding class, a Visual discriminator which is used to force each produced image to come from the same image space distribution as the given class and finally an Informative-negative Mining component whcih is used to actively seek regions of the latent space and force the generator to produce good in-class images even for these instances.

There are also alternative approaches that completely bypass the reconstruction step of AEs and focus more on the embedding component. One examples of this category of approaches is represented by the Geometric Transformation framework (GT) \cite{NEURIPS2018_5e62d03a} which proposes a model trained as a multi class classifier on different geometric transformations of the normal images, with the idea that by learning to discriminate between them the features learned by the model will be very ones which allow to efficiently detect novelties at test time.

A different idea is provided by Salehi et. al in \refcite{salehi2021multiresolution} where they train a cloner network that learns to mimic the comprehensive behaviour of a larger pre-trained expert network (e.g a VGG-16 network), by learning its knowledge regarding normal data and discarding all superfluous filters. Anomaly detection is then performed by analyzing the discrepancies in behaviour at test time.

Finally, we present VT-ADL\cite{mishra21-vt-adl} which, to our knowledge, is the only approach to this date focusing on the use of the ViT architecture to tackle the image anomaly detection problem. VT-ADL performs an embedding of each patch obtained from the original image into a latent space in which the positional information of the patches is also considered. These embeddings are then fed into two different modules, which allow the model to perform the anomaly detection and localization tasks. The first of the two modules is represented by a convolutional decoder which aims to reconstruct the original image starting from the patch embeddings while the second one is a Gaussian Mixture Density Network which is used to model the distribution of the normal data in the latent space in which the patches are embedded.

\begin{figure*}
    \centering
    \includegraphics[width=\linewidth]{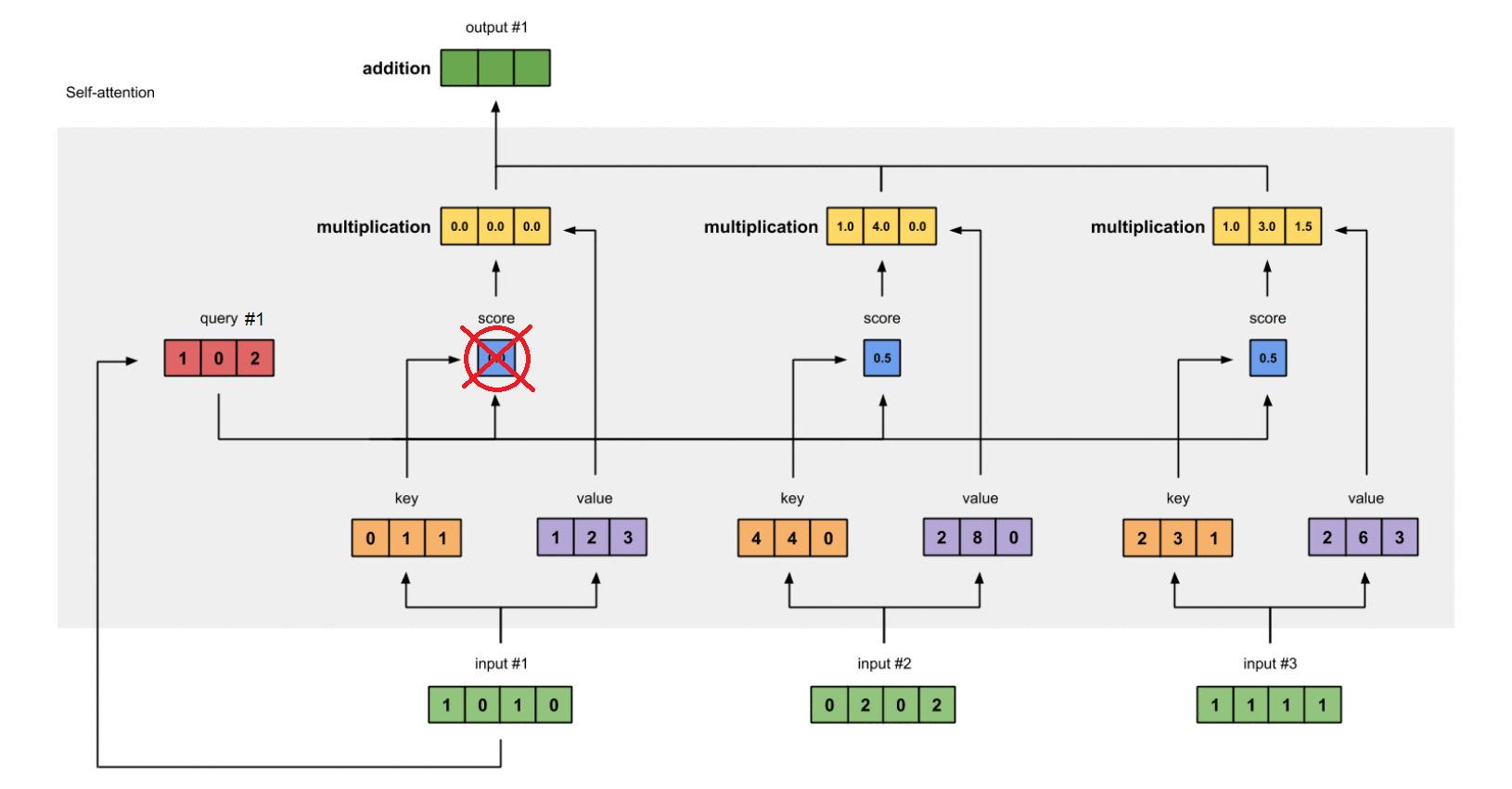}
    \caption{Illustration of the masking process performed in the self-attention module of our model. In the example we are calculating the attention values for patch \#1, therefore we set the dot product between Query \#1 and Key \#1 to 0 in order to take into account only the attention values of the remaining patches}
    \label{fig:masking}
\end{figure*}
\subsection{Vision Transformers}
The Vision Transformer is a deep neural network architecture proposed by Dosovitskiy et al.\cite{dosovitskiy2021an} as an alternative to convolutional-based architectures for computer vision applications. This model builds upon the idea of Self-Attention introduced in the original Transformer paper\cite{Vaswani2017AttentionIA}, which has since become the model of choice for Natural Language Processing (NLP) Applications, replacing Recurrent Neural Networks.

\subsubsection*{Architecture: }The Vision Transformer architecture (Fig.\ref{fig:vit}) differs from the original transformer one in the fact that it only uses the encoder module leaving out the Decoder. The encoder module, which takes as its input a set of flattened representations of the different patches composing the image we are trying to analyze, consists of a Stack of N identical Layers each containing two sub-layers: the first one is a multi-head self-attention block while the second one is a fully connected feed-forward layer. Around each of the 2 blocks, a residual connection is applied, followed by a layer normalization step.
The role of the encoder module in the ViT architecture is that of learning a correlation between the different patches composing an image.
\begin{figure*}
    \centering
    \includegraphics[width=\linewidth]{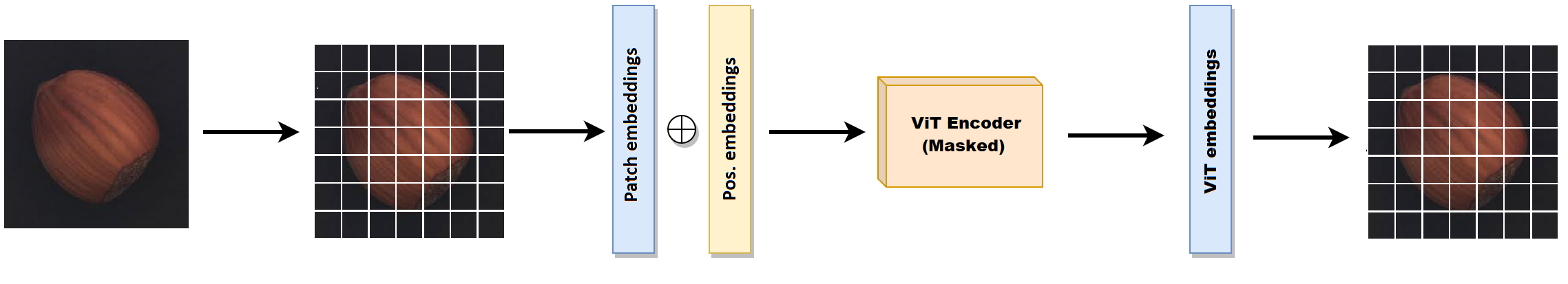}
    \caption{Overview of our model in its basic form: the original image is split into patches of the same size which is used as the input of a Multi-Layer Perceptron (MLP) in order to get the corresponding embedding. Then, to each patch embedding, a position embedding is added. The resulting tensor is then processed through a masked ViT encoder module in order to obtain the final embedding which will be used to reconstruct the respective patch by using an MLP Decoder}
    \label{fig:model1}
\end{figure*}
In order to preserve the spatial information regarding the position of the different patches a positional embedding is added to the patches representations before feeding them to the ViT encoder. The positional embedding can be fixed or learned alongside the other parameters.

\subsubsection*{Multi-head Attention: } The attention mechanism can be described as a function that maps triplets of vectors (represented by a query (Q) and key-value (K, V) pairs) to an output which is computed as a weighted sum of the values where the weights of each value are computed based on a compatibility relationship between a query and the corresponding key.
In Vision Transformers instead of performing the self-attention step only once for each set of queries, key and value, we project them in \textit{h} different spaces via learned linear projections. Attention is then calculated for each of these different projections and the final outputs are concatenated and projected again to obtain the final values. The whole self attention calculation process can be summarized by the following equation:

\begin{equation}
    Attention(Q,K,V) = softmax(\frac{QK^T}{\sqrt{d_k}})V
\end{equation}

This approach allows the model to consider different representation subspaces when calculating attention between the different parts of the image.

\section{Methods}
\label{sec:solution}
\subsection{Data preparation}
As far as data preparation is concerned we tried to adopt a very minimal approach. The only way the images themselves are pre-processed is through a resizing process that brings them all to the same final size, which is  128x128 for all the classes. This step was necessary in order to have a consistent number of patches when dividing the images in the next steps and it was also helpful for reducing the overall complexity of the model.
The only other operation applied to the dataset before training is a shuffling of the instances in order to avoid any potential bias for the model based on their order.
\subsection{Model description}
The model on which the work for this paper is based on the classical Vision transformer architecture presented in the previous section.

In this work, however, we introduce two very important changes the aformentioned architecture which we will describe in detail in the present section. The first change is represented by a masking process by which we try to remove the focus of each patch of the original image on its own features, redirecting the attention to the remaining ones. While the second idea we introduced regards the division of the original images in multi shaped and multi scale patches, as opposed to relying only on the traditional square patches presented in the original work.

\subsubsection{Masking process}
While many of the image-reconstruction-based models for anomaly detection work on the entire image, this approach leads to a very common problem. Since the model uses all the information available in the input it tends to learn how to reconstruct correctly the anomalous images as well as the normal one which, of course, is not the desired behavior since the goal is to discriminate between the two classes.
For this reason, the idea behind our proposed method is to mask out some of the information of the original image and use only the remaining data to perform the reconstruction. This as we will show, greatly reduces the problem mentioned in the previous paragraph. More specifically we decided to work on patches representing non-overlapping subsets of the original image and to reconstruct each of these patches based only on the content of the remaining ones. For this purpose, we decided to adopt the ViT architecture as a baseline for our model, which as we have previously shown, allows finding a correlation between different parts of an image by calculating an attention score between its patches.
In its basic form, though, the ViT model provides for each patch an embedding obtained by processing the whole information of the original image, including the patch itself. For this reason, we altered the concept of self-attention introduced by the original model in order to mask out this piece of information. In order to do so, we added a masking module in the multi head self attention component of the original architecture which forces the dot product between the key generated for each patch and the respective query to be set to 0, as shown in Fig.~\ref{fig:masking}.
In other words, after obtaining the $nxn$ (where n is the number of patches) matrix in which the cell in position $ij$ represents the correlation between patches i and j, we set its diagonal, representing the internal information of each patch, to 0 in order to use only external information for its reconstruction. \footnotetext{Source code will be available on Github after manuscript acceptance}
\begin{figure*}[htb]
    \includegraphics[width=\linewidth]{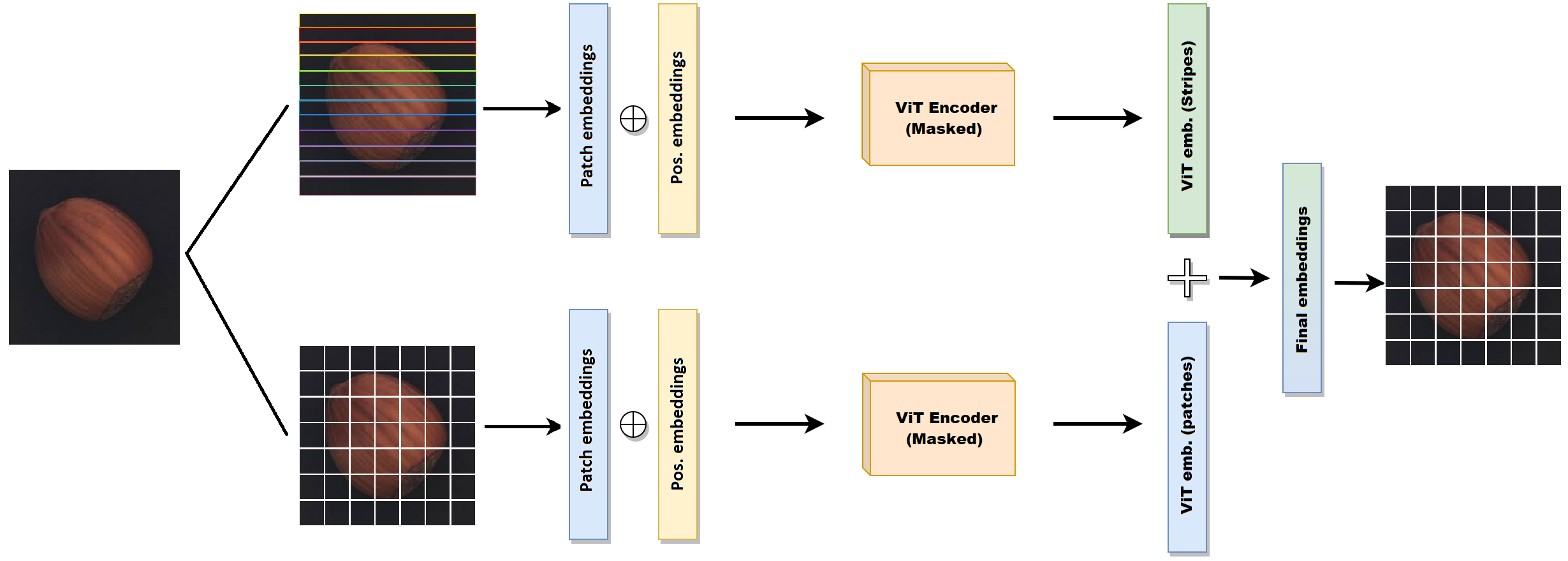}
    \caption[]{Overview of the proposed model using horizontal stripes only, in addition to the square patches\footnotemark}
    \label{fig:ourmodel}
\end{figure*}
An overview of the model described so far is given in Fig.~\ref{fig:model1}.

\begin{figurehere}
    \centering
    \includegraphics[width=\linewidth]{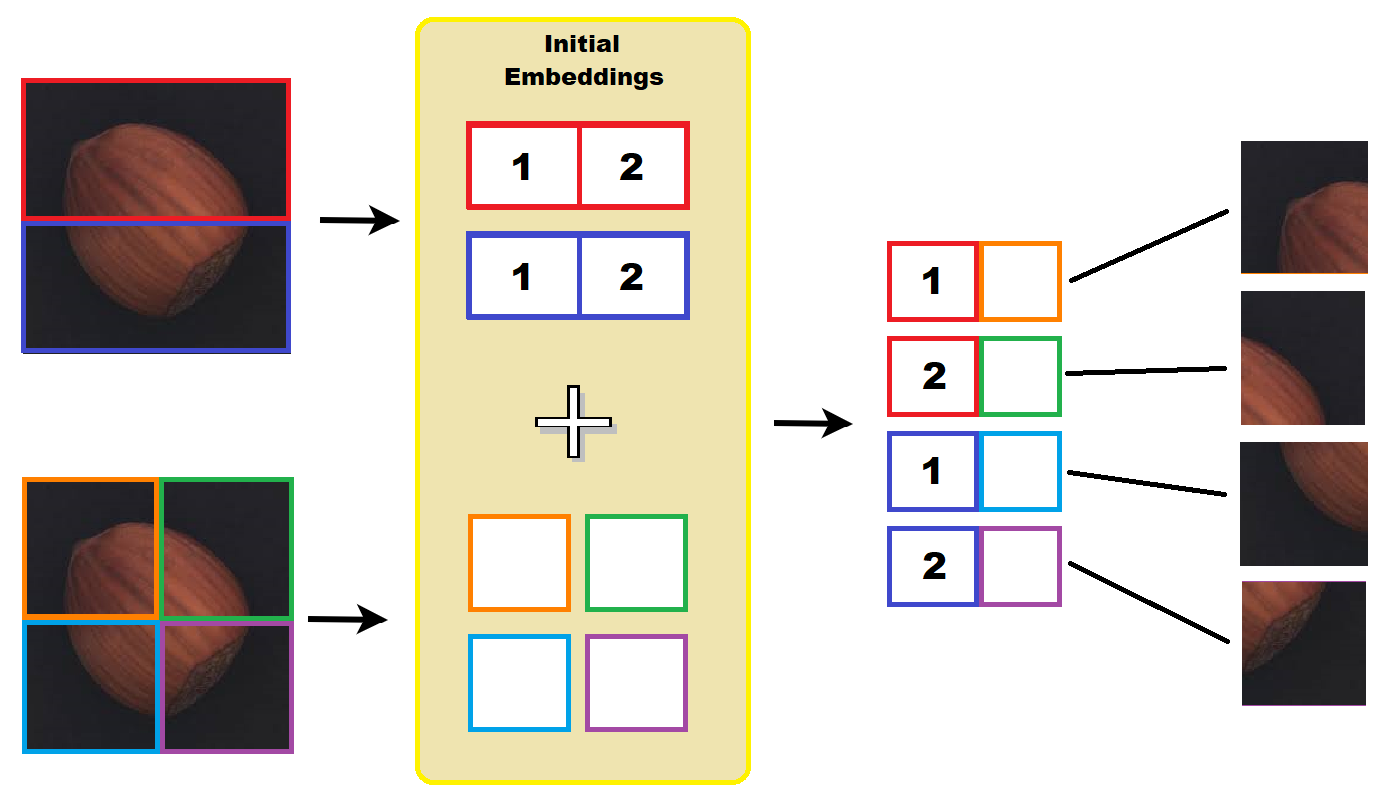}
    \caption{Illustration of the process carried out for the concatenation of the patch embeddings. Two components are concatenated together, the first one is the embedding obtained trough the self attention process performed on square patches, while the second one is a subset of the embedding resulting from the self attention process carried out on strip patches.}
    \label{fig:embconcat}
\end{figurehere}

\subsubsection{Multi shape patch structure}
Another direction in which we expanded the original idea of the ViT model regards the way the patches fed to it are obtained from the original image. While the use of square patches is a common choice, it is ultimately an arbitrary one, for this reason, we introduced in our model a set of patches with different shapes, in particular horizontal and vertical stripes, each of which was processed in the same way as the square ones. The encodings of the different types of patches have then been concatenated to form the final representation for each patch which has then been used for its reconstruction. The reconstruction process involved the use of a simple MLP Decoder which worked on each patch embedding singularly to reproduce the respective patch. This process has been carried out by ensuring that the masking property of the network was held true, to do so the concatenation between patches of different shapes has been carried on based on their spatial location in the original image so that each square patch was fully contained in the respective horizontal (or vertical) stripe patch.
Furthermore, in order to keep the final embedding size as small as possible, we split the embedding of each stripe into $p$ segments of the same size, where $p=N/K$ (with N=size of the image and K=size of the square patches) is the number of square patches contained in each stripe, and then concatenated each of these segments to the embedding of a different patch thus forcing the model to learn specific information about each of them in different locations of the stripe embedding (Fig.~\ref{fig:embconcat}).
A high level overview of our final model is provided in Fig.~\ref{fig:ourmodel}.
\section{Evaluation}
\label{sec:evaluation}
\subsection{datasets}
To test our Framework we relied on two popular dataset, namely MVTec which is currently one of the most popular and heterogeneous datasets in the context of anomaly detection for industrial quality control, and HeadCT which is instead a dataset representing a collection of X-Ray head scans in which the anomalies are represented by the presence of hemorrhages. While the use of larger datasets is usually preferable, the options available in the context of anomaly detection are still very limited. MVTec already represents a big step forward when it comes to size and heterogeneity of datasets in this area of research compared to the ones that were available before it and, combined with the HeadCT dataset, we believe they provide and reliable option for the evaluation of the proposed approach.

\begin{figurehere}
    \begingroup
    \centering
    \includegraphics[width=.7\linewidth]{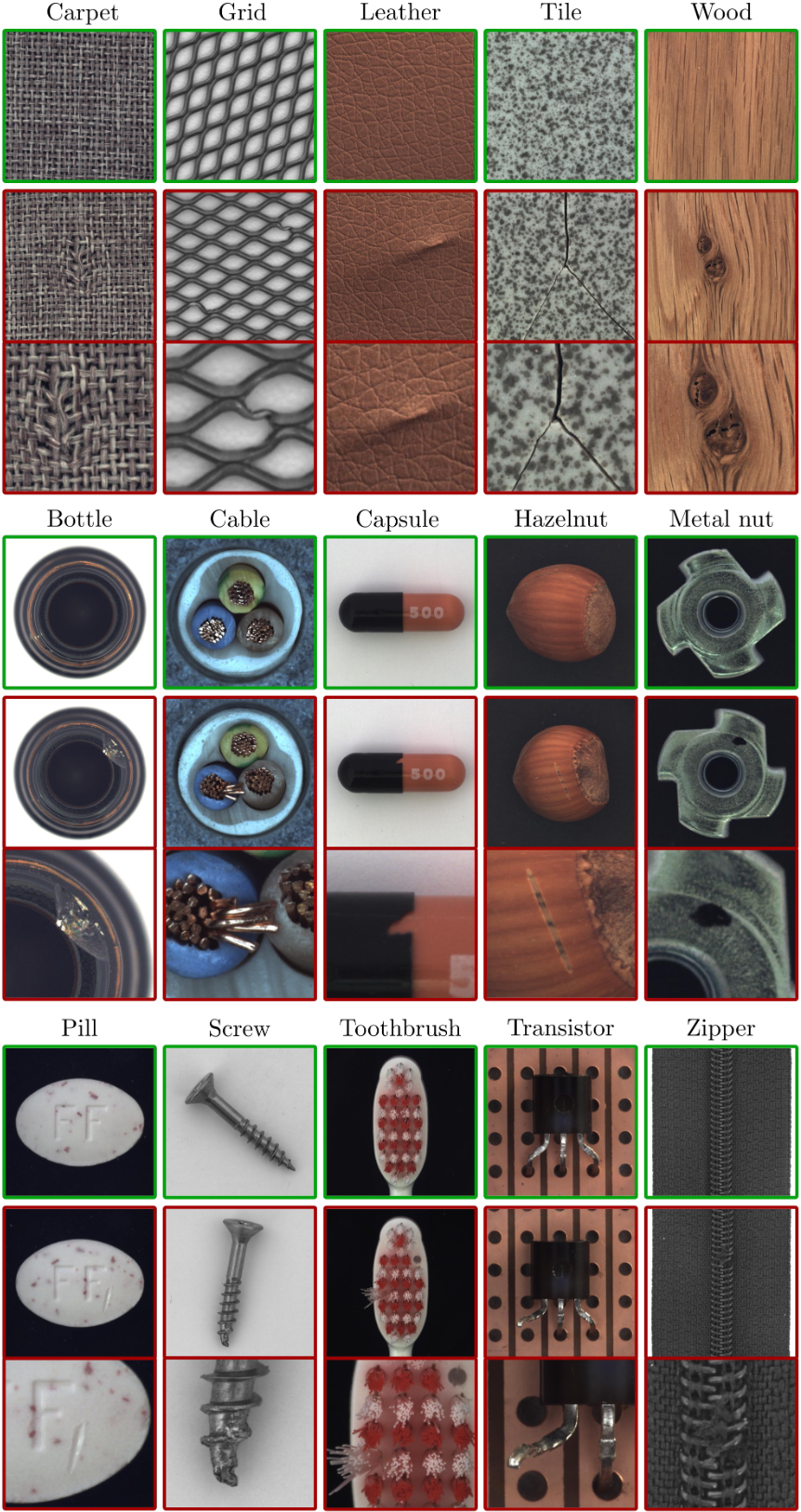}
    \caption{Sample for each of the classes present in the MVTec dataset. For each of them is provided a normal instance (top row), an instance containing an anomaly (middle row), and a zoomed view on the defect (bottom row)}
    \label{fig:mvtec}
    \endgroup
\end{figurehere}
\subsubsection{The MVTec Dataset}
The dataset we used for both the training and testing of our model is the MVTec Anomaly Detection dataset\cite{Mvtec2019}, which has become a common benchmark in recent works on anomaly detection and localization. This dataset consists of 3629 training images and 1725 testing images divided into 15 classes, five of which represent different textures and the remaining ones covering a set of products with heterogeneous characteristics, some of them present a rigid structure, others are deformable or present natural variations in their appearance. Furthermore, the way in which the images are captured is also heterogeneous, for some of the classes all the instances belonging to the present the product in a roughly aligned fashion while for some others a random rotation is introduced. Since grayscale images are also a common occurrence in industrial settings three classes are made available only as single-channel images. The testing set is also highly heterogeneous in its composition, providing a wide variety of anomalies. In total 73 different types of defects are provided, 5 for each category on mean. To make the testing procedure possible pixel-accurate labels for all the defective regions in each image are also provided by the authors.
An overview of the different classes of textures and products in the dataset, together with an anomalous example for each of them is provided in Fig.~\ref{fig:mvtec}.
Furthermore, in Table~\ref{tab:mvtec} the details regarding each class are reported. As we can notice there are no anomalous instances in the training set, this because usually the anomaly detection models are trained only on normal data.

\begin{table*}[htb]
\centering
\tbl{MVTec dataset details\label{tab:mvtec}}{	
\begin{tabular}{|l|ccccc|}
\hline
 &
  \multicolumn{1}{c|}{\textbf{Class}} &
  \multicolumn{1}{c|}{\textbf{\begin{tabular}[c]{@{}c@{}}Train\\ (Normal)\end{tabular}}} &
  \multicolumn{1}{c|}{\textbf{\begin{tabular}[c]{@{}c@{}}Test\\ (Normal)\end{tabular}}} &
  \multicolumn{1}{c|}{\textbf{\begin{tabular}[c]{@{}c@{}}Test\\ (Anomaly)\end{tabular}}} &
  \textbf{Image side} \\ \hline
\multicolumn{1}{|c|}{\multirow{5}{*}{\textbf{Textures}}} & \textbf{Carpet}     & 280 & 28 & 80  & 1024 \\
\multicolumn{1}{|c|}{}                                   & \textbf{Grid}       & 264 & 21 & 57  & 1024 \\
\multicolumn{1}{|c|}{}                                   & \textbf{Leather}    & 245 & 32 & 92  & 1024 \\
\multicolumn{1}{|c|}{}                                   & \textbf{Tile}       & 230 & 33 & 84  & 1024 \\
\multicolumn{1}{|c|}{}                                   & \textbf{Wood}       & 247 & 19 & 60  & 1024 \\ \hline
\multirow{10}{*}{\textbf{Products}}                      & \textbf{Bottle}     & 209 & 20 & 63  & 900  \\
                                                         & \textbf{Cable}      & 224 & 58 & 92  & 1024 \\
                                                         & \textbf{Capsule}    & 219 & 23 & 109 & 1000 \\
                                                         & \textbf{Hazelnut}   & 391 & 40 & 70  & 1024 \\
                                                         & \textbf{Metal nut}  & 220 & 22 & 93  & 700  \\
                                                         & \textbf{Pill}       & 267 & 26 & 141 & 800  \\
                                                         & \textbf{Screw}      & 320 & 41 & 119 & 1024 \\
                                                         & \textbf{Toothbrush} & 60  & 12 & 30  & 1024 \\
                                                         & \textbf{Transistor} & 213 & 60 & 40  & 1024 \\
                                                         & \textbf{Zipper}     & 240 & 32 & 119 & 1024 \\ \hline
\end{tabular}
}
\end{table*}

\subsubsection{Head CT Dataset}
As a further benchmark for our model, we used the Head CT dataset\cite{headct}, which we selected because of its medical nature and because of its relatively small size, which allowed us to prove the potential of the Vision Transformer architecture even in this particularly challenging type of setting.
The head ct dataset is, in fact, composed of a total of 200 images 100 of which represent head ct of healthy individuals while the remaining 100 represent scans of patients with a head hemorrhage.
For the purpose of this study, we adopted 80 normal images for the training of the model, while the remaining 120 instances (20 normal, 100 with an hemorrhage) were used in the testing process.
Some samples representing normal and anomalous images from the dataset are reported in Fig.~\ref{fig:headct}.

\begin{figurehere}
    \centering
    \includegraphics[width=\linewidth]{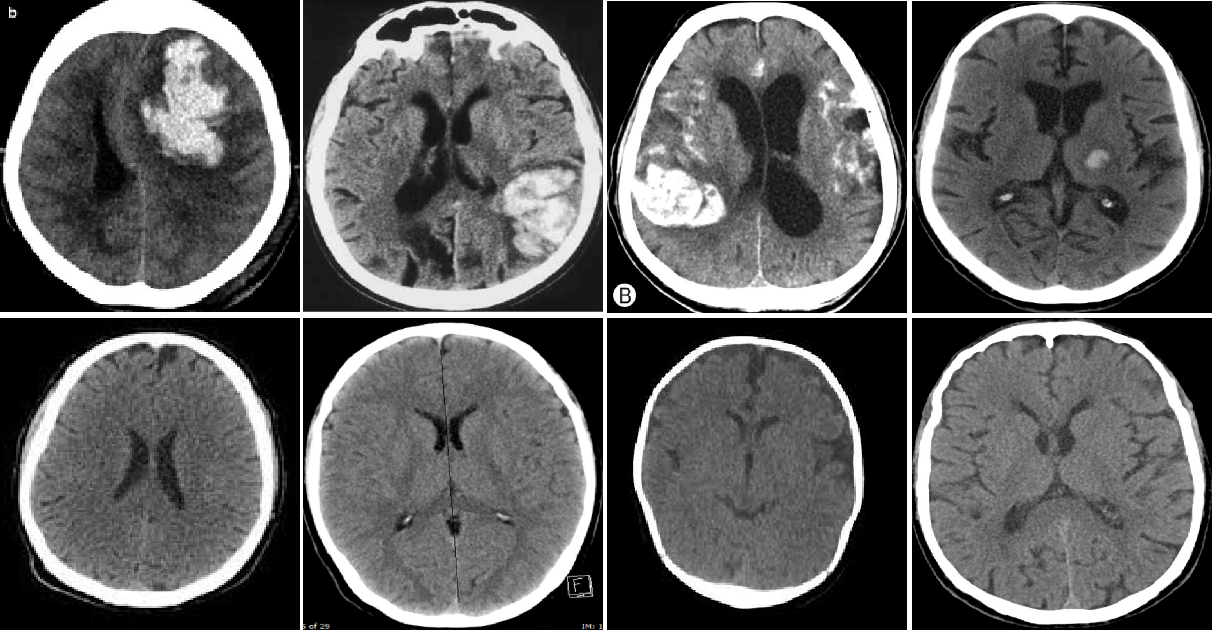}
    \caption{Samples of anomalous (top row) and normal (bottom row) images from the Head CT dataset}
    \label{fig:headct}
\end{figurehere}

\subsection{Training setup}
The training of the model was performed in a self-supervised fashion, using only the normal images of the dataset, over a maximum of 3000 epochs with an early stop introduced after epoch 500 which would trigger when the performance of the model on the validation set didn't improve in the previous 50 epochs. 
The validation set was composed of 10\% of the images present in the training data of the datasets. While often larger percentages are used, both MVTec and HeadCT are relatively small datasets, which led us to prefer keeping as much data as possible for the training of the model.
The total loss function we used is obtained by summing the L1 loss and the negative of the SSIM Similarity, as defined in [\refcite{SSIM2004}], calculated between the original image given as input and the reconstructed image returned as the output of the model. A formal description of the two functions is given hereafter:

\begin{gather}
L_1(X,\hat{X})=\sum_{i=0}^{h-1}\sum_{j=0}^{w-1}{|X_{ij}-\hat{X}_{ij}|}\\
L_{SSIM}(X,\hat{X})= -\frac{(2\mu_{X}\mu_{\hat{X}} +c_1)(2\sigma_{X\hat{X}}+c_2)}{(\mu_X^2+\mu_{\hat{X}}^2+c_1)(\sigma_X^2+\sigma_{\hat{X}}^2+c_2)}\\
L(X,\hat{X}) = L_1(X,\hat{X}) + L_{SSIM}(X,\hat{X})
\end{gather}

Where:
\begin{itemize}
    \item \textbf{X: } Is the original image
    \item $\boldsymbol{\hat{X}}$: Is the reconstructed image
    \item \textbf{h, w: } Are the height and width of the image in pixel
    \item $\boldsymbol{\mu_{x}}$: Is mean value of image x
    \item $\boldsymbol{\sigma_{x}^2}$: Is variance of image x
    \item $\boldsymbol{\sigma_{xy}}$: Is the covariance of x and y
    \item $\boldsymbol{c_1=(k_1L)^2 \& c_2=(k_2L)^2: }$ Are two variables used to stabilize the division with weak denominator
    \item \textbf{L: } Is the dynamic range of the pixel-values (usually $2^{\# bits per pixel}-1)$
    \item $\boldsymbol{k_1 \& k_2}$: are two costants set to 0.01 and 0.03 respectively.
\end{itemize}
It is important to notice that the SSIM function represents a similarity measure defined in the interval $[-1,1]$ where 1 means that the two images being compared are identical. For this reason, to use it as a loss function it needs to be negated.

The use of an $L_1$ loss instead of a more classical MSE loss is motivated by the fact that it reduces blurriness and color artifacts in the reconstructed images\cite{ZhaoNvidia2017}.

The hyperparameters adopted during the training process are shown in Table~\ref{tab:hyperparams}. We tried to keep the Structure of the TF encoders as shallow as possible and actually noticed that increasing the number of blocks improved the performance of the model only marginally for the texture classes while for the product classes didn't help at all. A possible reason for this is that by increasing the depth, the model would become too complex for the relatively small dataset we used for training and evaluation and thus lead to overfitting. 

Another important hyperparameter that needs to be selected when adopting a ViT architecture is the size of the patches in which the image needs to be divided. Following the original ViT paper with adopted 16x16 patches as the baseline for our model, which empirical tests confirmed to be a proper value for the task considered. In general the idea to keep in mind during patch size selection is that Vision Transformers tend to work better with a high number of sequences as their input, for this reason the size of the patches needs to be kept relatively small compared to the image size. Furthermore in the context of the proposed approach it was important not to use patches that are too small, as this would make much more frequent the presence of anomalies crossing multiple patches, which are easier to reconstruct since they can be inferred from the surroundings of the current patch analyzed and, therefore are more difficult to detect. The size of the short side of the stripes was selected to match the square patches sizes, while the long side is determined by the image size.

\begin{table*}[htb]
\centering
\tbl{model hyperparameters\label{tab:hyperparams}}{
\resizebox{\textwidth}{!}{\begin{tabular}{|l|c|c|c|c|c|c|c|c|}
\hline
   \textbf{Classes} & \textbf{LR} & \textbf{Batch} & \textbf{Img size} & \textbf{Patch size}& \textbf{Stripe size}& \textbf{Emb. size}& \textbf{\#Heads} & \textbf{\#Blocks}\\ \hline
\textbf{Textures} & 1e\textasciicircum{}-4 & 64               & 128x128                & 16x16  & 128x16& 128 & 4 & 2                                           \\ \hline
\textbf{Products}  & 1e\textasciicircum{}-4 & 64               & 128x128                 & 16x16  & 128x16& 128    & 4 & 1                                         \\ \hline

\end{tabular}}}
\end{table*}

\begin{figurehere}
    \centering
    \includegraphics[width=0.9\linewidth]{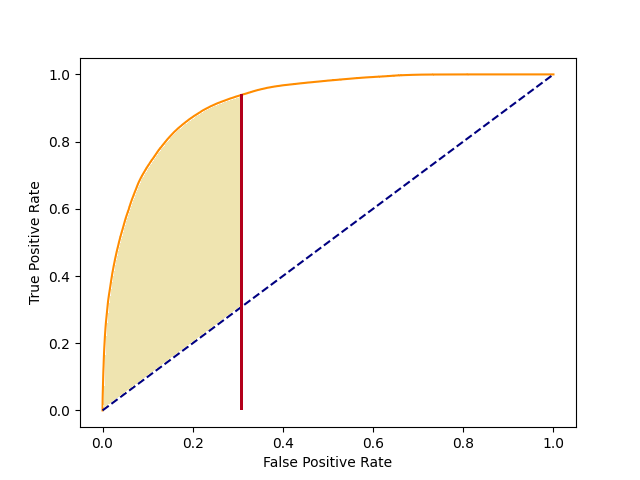}
    \caption{Illustration of the thresholded AUROC metric. The red line represents the selected threshold while the yellow region is the area taken in consideration for the evaluation of the model.}
    \label{fig:rocurve}
\end{figurehere}

\subsection{Metrics}
For the evaluation of our model, since our objective is to assess its anomaly segmentation capabilities, we opted to use the Area Under the ROC Curve (AUROC), which plots the False Positive Rate versus the True Positive Rate and is a very commonly used metric for this type of problems. In order to obtain consistent results with those presented by the authors of the MVTec dataset, the values of the metric are computed up to a False Positive Rate of 0.3. The reason behind this choice is that thresholds that yield to a high FPR lead to meaningless segmentation results, especially for industrial scenarios where such results would lead to wrongly rejecting products not presenting any defects. An illustration of the thresholding process is provided in Fig.\ref{fig:rocurve} 

\subsection{Results}
All the values presented in this section for our models have been obtained by first generating the reconstructed images, patch by patch, for each of the instances in the test set of the MVTec dataset, each of them has then been compared to the respective original image by applying a pixel-wise MSE loss (the deltas over the three channels of each pixel have been summed), in order to obtain a heatmap that highlighted the largest differences between the two, thus revealing the potentially  anomalous regions. A Gaussian filter has also been applied to the map to smooth out anomalous regions and reduce noise.

In Table~\ref{tab:ablation} the results for our ablation study, in which we compare the performance of our model when applying the masking process and different combinations of patch shapes, are reported. As we can see the approach which relies solely on the traditional square patches is outperformed by every other method using a combination of different shapes. In particular, we show that the best performing approach is the one represented by a combination of the square patches with horizontal, or vertical stripes, which leads to an improvement in the pixelwise AUROC value of more than 4\% compared to the baseline approach. As we can see the orientation of the stripes combined with the square patches didn't affect the overall performance of the model, which achieved, on mean, the same performance when using horizontal or vertical ones. On the other hand, we can see how using both orientations together leads to a decrease in the performance of the model. Our guess is that in this scenario the model becomes too complex compared to the relatively small dataset we used and therefore becomes too specialized on the training set, leading to overfitting. Finally, we can notice how completely removing the square patches and relying only on the horizontal and vertical stripes also affects the model performance negatively. A possible explanation for this behavior is that the square patches are able to provide some more locally specific information that the model needs to perform the reconstructions of the single patches effectively.
\begin{table*}[htb]

\tbl{results of the ablation study in which we show the effect of using different combinations of patch shapes on the model performance. All the values reported representing the normalized area under the ROC curve up to an mean FPR per-pixel of 30\%.\label{tab:ablation}}{
\centering
\begin{tabular}{lllllll}
\hline
\multicolumn{1}{c}{Class} &
\multicolumn{1}{c}{\begin{tabular}[c]{@{}c@{}}Only Squares \\ (No Mask)\end{tabular}} &
  \multicolumn{1}{c}{\begin{tabular}[c]{@{}c@{}}Only Squares \\ (Masked)\end{tabular}} &
  \multicolumn{1}{c}{Squares + Rows} &
  \multicolumn{1}{c}{Squares + Cols} &
  \multicolumn{1}{c}{Rows + Cols} &
  \multicolumn{1}{c}{\begin{tabular}[c]{@{}c@{}}Squares + \\ Rows + \\ Cols\end{tabular}} \\ \hline
Carpet     & 0.495 &0.510    & 0.712         & 0.723          & \textbf{0.755} & 0.661 \\
Grid       & 0.814 &0.835   & \textbf{0.884} & 0.883          & 0.800          & 0.817 \\
Leather    & 0.734 &0.792   & \textbf{0.976} & 0.905          & 0.723          & 0.723 \\
Tile       & 0.727 &0.754   & 0.771          & \textbf{0.772} & 0.659          & 0.690 \\
Wood       & 0.701 &0.757   & 0.836          & \textbf{0.851} & 0.808          & 0.840 \\
Bottle     & 0.796 &0.812   & \textbf{0.850} & 0.847          & 0.821          & 0.842 \\
Cable      & 0.683 &0.715   & 0.701          & 0.701          & \textbf{0.753} & 0.700 \\
Capsule    & 0.863 &\textbf{0.895}   & 0.891          & 0.885          & 0.889          & 0.893 \\
Hazelnut   & 0.927 &0.951   & \textbf{0.953} & 0.945          & 0.951          & 0.942 \\
Metal nut  & 0.712 &0.721   & 0.773          & 0.779          & \textbf{0.865} & 0.806 \\
Pill       & 0.815 &0.835   & 0.852          & \textbf{0.852} & 0.806          & 0.850 \\
Screw      & 0.770 &0.818   & 0.901          & \textbf{0.905} & 0.903          & 0.903 \\
Toothbrush & 0.926 &0.949   & \textbf{0.975} & 0.966          & 0.967          & 0.971 \\
Transistor & 0.848 &0.855   & 0.841          & 0.860          & \textbf{0.866} & 0.865 \\
Zipper     & 0.783 &\textbf{0.805}   & 0.750          & 0.770          & 0.634          & 0.720 \\ \hline
Mean       & 0.773 &0.8     & \textbf{0.844} & \textbf{0.843} & 0.814          & 0.815 \\ \hline
\end{tabular}}
\end{table*}

In Table~\ref{tab:comparison} we provide a comparison between our methods and other approaches. In particular, we focus on the methods proposed in the MVTec paper\cite{bergmann2021mvtec} as our benchmarks and on VT-ADL as the only other approach using a ViT architecture for anomaly localization. For the latter, the results shown have been calculated by us as the original paper didn't provide the values for the AUROC metric, while for every other model the result has been gathered from the original paper.
As we can see our approach vastly outperforms the previous method based on ViTs represented by VT-ADL, in particular, by referring back to Table~\ref{tab:ablation}, we can see that even the approach relying solely on square patches still achieves better results than VT-ADL by improving over its results by almost an 12\% margin, therefore proving the effectiveness of the masking process introduced in the Self-attention module of our model.
 \begin{table*}[htb]
 \tbl{Normalized area under the ROC curve up to an mean false positive rate per-pixel of 30\% for each dataset category. The values in bold represent the best scores overall, while the underlined ones represent the best scores between the models not using extra data.\label{tab:comparison}}{
\resizebox{\textwidth}{!}{\begin{tabular}{llllllllll}
\hline
\multicolumn{1}{c}{Class} &
  \multicolumn{1}{c}{f-anoGan} &
  \multicolumn{1}{c}{Feature dictionary} &
  \multicolumn{1}{c}{Student teacher} &
  \multicolumn{1}{c}{l2-AE} &
  \multicolumn{1}{c}{SSIM-AE} &
  \multicolumn{1}{c}{Texture inspection} &
  \multicolumn{1}{c}{Variation model} &
  \multicolumn{1}{c}{VT-ADL} &
  \multicolumn{1}{c}{Ours} \\ \hline
Carpet     & 0.251 & \textbf{0.943} & 0.927          & 0.287 & 0.365 & \underline{0.874} & 0.162 & 0.549 & 0.712         \\
Grid       & 0.550 & 0.872          & \textbf{0.974} & 0.741 & 0.820 & 0.878 & 0.488 & 0.569 & \underline{0.884}          \\
Leather    & 0.574 & 0.819          & \textbf{0.976}          & 0.491 & 0.356 & 0.975 & 0.381 & 0.817 & \underline{\textbf{0.976}} \\
Tile       & 0.180 & 0.854          & \textbf{0.946} & 0.174 & 0.156 & 0.314 & 0.304 & 0.589 & \underline{0.771}          \\
Wood       & 0.392 & 0.720          & \textbf{0.895}         & 0.417 & 0.404 & 0.723 & 0.408 & 0.682 & \underline{0.836} \\
Bottle     & 0.422 & \textbf{0.953} & 0.943          & 0.528 & 0.624 & 0.454 & 0.667 & 0.687 & \underline{0.850}          \\
Cable      & 0.453 & 0.797          & \textbf{0.866} & 0.510 & 0.302 & 0.512 & 0.423 & 0.751 & 0.701          \\
Capsule    & 0.362 & 0.793          & \textbf{0.952} & 0.732 & 0.799 & 0.698 & 0.843 & 0.615 & \underline{0.891}         \\
    Hazelnut   & 0.825 & 0.911          & \textbf{0.959}        & 0.879 & 0.847 & 0.955 & 0.802 & 0.926 & \underline{\textbf{0.959}} \\
Metal nut  & 0.435 & 0.862          & \textbf{0.979} & 0.572 & 0.539 & 0.135 & 0.462 & 0.711 & \underline{0.773}          \\
Pill       & 0.504 & 0.911          & \textbf{0.955} & 0.690 & 0.698 & 0.440 & 0.666 & 0.748 & \underline{0.852}          \\
Screw      & 0.814 & 0.738          & \textbf{0.961} & 0.867 & 0.885 & 0.877 & 0.697 & 0.771 & \underline{0.901}          \\
Toothbrush & 0.749 & 0.916          & 0.971          & 0.837 & 0.846 & 0.712 & 0.775 & 0.878 & \underline{\textbf{0.975}} \\
Transistor & 0.372 & 0.527          & 0.566          & 0.657 & 0.562 & 0.363 & 0.601 & 0.689 & \underline{\textbf{0.860}} \\
Zipper     & 0.201 & 0.921          & \textbf{0.964} & 0.474 & 0.564 & \underline{0.928} & 0.209 & 0.683 & 0.750          \\ \hline
Mean       & 0.472 & 0.836          & \textbf{0.922} & 0.590 & 0.584 & 0.656 & 0.526 & 0.683 & \underline{0.844}          \\ \hline
\end{tabular}}}
\end{table*}
As for the remaining models, we show that while our approach performs worst than the best one from the MVTec paper, represented by the student-teacher architecture (7.8\% AUROC score difference), it outperforms every other method by a margin going from 1\%  to 37\%. One important aspect to notice is that the two top-performing methods presented in the MVTec paper, namely the student-teacher and the Feature Dictionary models which are the only ones exceeding an mean AUROC score of $0.7$, both rely on feature extractors pre-trained on much larger datasets such as the popular imagenet one\cite{deng2009imagenet} while our model is trained from scratch on the MVTec dataset making it the best performing model not relying on extra data for training and thus showing the possibility of adopting transformer-based models, typically considered very heavy, even to scenarios where we have a relatively small amount of data available.

\begin{figure*}
    \centering
    \includegraphics[width=0.9\linewidth]{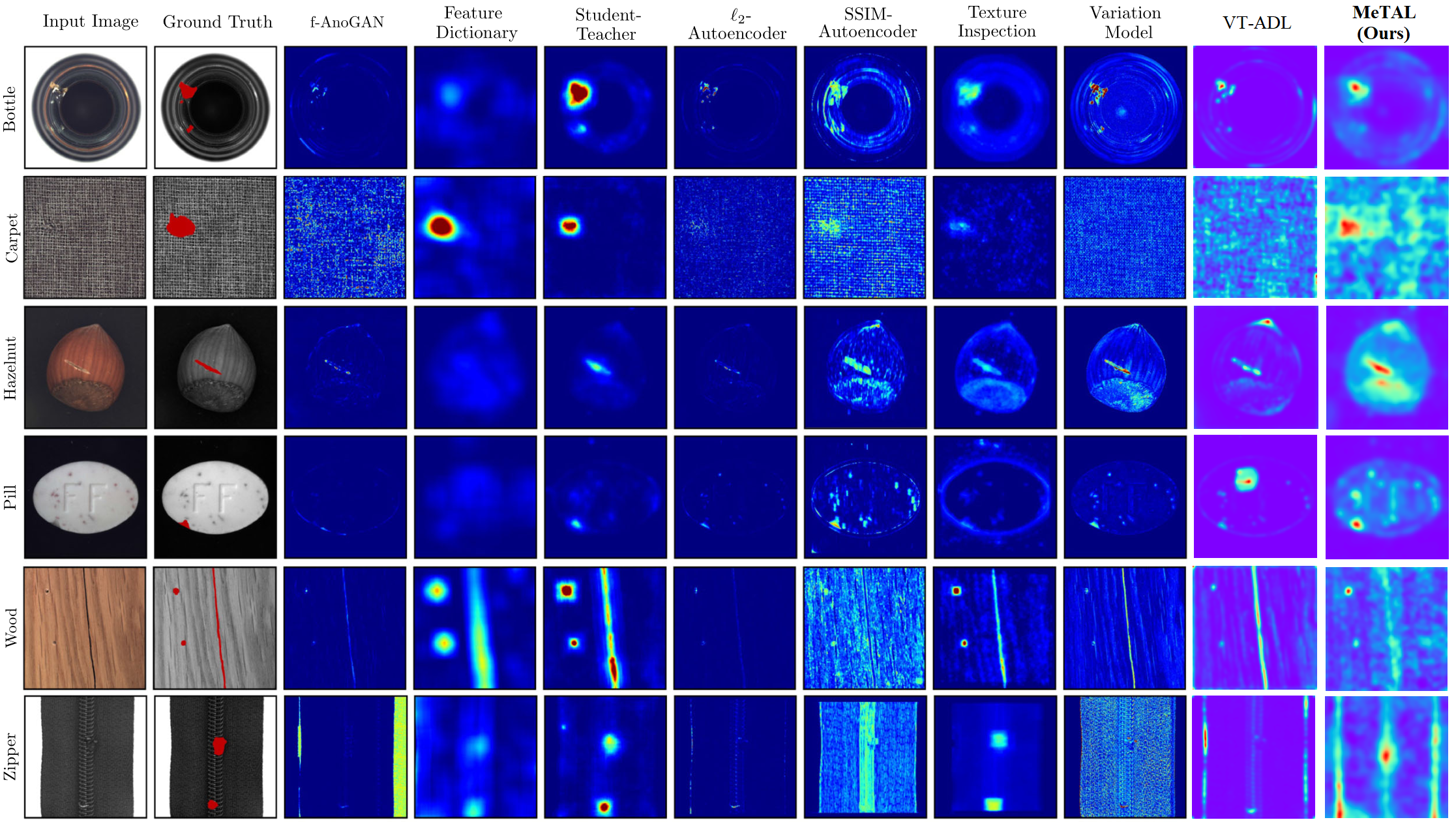}
    \caption{In order from left to right, we have: (1) The original image used as input, (2) the ground truth mask showing the location of the anomaly in the original image, and (3-11) the anomaly maps generated by our model and the competition}
    \label{fig:anomaps}
\end{figure*}

Finally, in Fig.~\ref{fig:anomaps}, we provide some qualitative results of our model by showing a comparison between the original images and ground truth anomaly maps with the reconstructed images and anomaly maps generated by our model. As we can see the model is able to effectively mask out the smaller anomalies from the reconstructed images, leaving only small artifacts in their places. As for larger anomalies that are spread through different regions of the original images, the model is usually not able to completely remove them as it can infer their structure from the surrounding patches. Nonetheless, in many cases, the defective part has a more ``washed out'' appearance in the reconstructed image which allows for its localization.

Furthermore, we present the results for the Head CT - hemorrhage dataset. In particular, since this dataset doesn't provide masks for the instances anomalies, we use as a comparison metric the images ROCAUC score.
We report our results, together with the ones presented by Salehi et al.\cite{salehi2021multiresolution} which, as far as we know, are the best ones available online for the headct dataset, in Tab.\ref{tab:headct_results}. 
As we can see our model, even in its most basic configuration, achieves results comparable to the State of the Art, while when horizontal stripes are added it is able to surpass the other presented approaches by a significant margin showing the effectiveness of the presented approach even for scenarios when the data available is very limited.
The downside, however, is that the increased complexity of the model appears to make it less stable as we can observe from the higher variance characterizing the results achieved by this configuration.

\begin{tablehere}
\tbl{AUROC for anomaly detection on Head CT dataset\label{tab:headct_results}. }{
\centering
\resizebox{\columnwidth}{!}{
\begin{tabular}{|l|c|c|c|c|c|c|c|c|}
\hline
   \textbf{} & \textbf{AUROC Head CT (\%)} \\   \hline
   \textbf{OCGAN \cite{Perera2019OCGANON}} & $51.20 \pm 0.358$\\ \hline
\textbf{LSA \cite{Abati2019LatentSA}} & $81.67 \pm 3.626$\\ \hline
\textbf{GT \cite{NEURIPS2018_5e62d03a}}  & $ 49.5 \pm 3.873$  \\ \hline
\textbf{MKD \cite{salehi2021multiresolution}}  & $80.4 \pm 0.006 $  \\ \hline \hline
\textbf{MeTAL (Squares Only)}  & $81.35 \pm 1.153$   \\ \hline
\textbf{MeTAL (Squares+Rows)}  & \textbf{86.32} $\pm 4.03$  \\ \hline  
\end{tabular}}}
\end{tablehere}
\begin{tablehere}
\tbl{Comparison of the number of parameters defining the compared models architectures\label{tab:models_params}}{
\centering
\begin{tabular}{|l|c|c|c|c|c|c|c|c|}
\hline
   \textbf{} & \textbf{\# of parameters} \\   \hline
   \textbf{$l^2$ - AE / SSIM-AE} & 1.2M\\ \hline
   \textbf{f-AnoGAN} & 24.6M\\ \hline
   \textbf{Feature dictionary} & 11.5M\\ \hline
\textbf{Student teacher} & 26M\\ \hline
\textbf{VT-ADL}  & 25M \\ \hline
\textbf{MKD}  & 15M  \\ \hline \hline
\textbf{Ours (Products)}  & 17.8M    \\ \hline
\textbf{Ours (textures)}  & 26.3M    \\ \hline 
\end{tabular}}
\end{tablehere}

For completeness we are also providing a comparison of the model sizes for the different approaches, expressed in number of parameters defining their structure (Tab. \ref{tab:models_params}). 
As we can see the proposed approach size is generally comparable to the ones of the other frameworks, excluding the most simple ones being represented by the two autoencoders variations which, however, achieve very poor performances on the selected dataset.

\section{Conclusions and Future Work}\label{sec:concl}
In the present paper, we investigated the use of a Framework based on the Self-Attention mechanism introduced by the Vision Transformer model. In particular, we proposed an image Inpainting approach to anomaly detection which leverages the ability of the ViT encoder to find correlations between different regions of a given image in order to reconstruct each of them based only on the information contained in the surrounding ones. Furthermore, we have shown that the model's performance is affected positively by the use of heterogeneous shapes for the subsets into which the original image is split. Our opinion is that this approach allows the model to learn correlations between patches at different scales of the image, therefore increasing the quality of the generated reconstructions. Finally, we have shown how the ViT model, while usually considered a heavy architecture, can be used effectively even on a relatively small dataset without the need to rely on extra data from external sources, achieving the best performance compared to other model's trained in a similar setting on the MVTec dataset.

Nonetheless, as future work, we believe it would be interesting to explore our model's capabilities when pre-trained on a larger dataset before fine-tuning it on the final task's data and we would also like to investigate the possibility of a more general approach to the multi-scale patch acquisition we introduced in this work.

As another line of reasearch we believe it would be also worth investigating the adoption of ideas introduced in other recent works present in the literature, such as [\refcite{Ahmadlou2010}, \refcite{pereiraFEMa}, \refcite{Alam2020}, \refcite{Rafiei2017}], in order to try to further improve the performance of the proposed architecture.

Furthermore, we believe that the main limitation of the proposed approach regards the ability to handle large scale anomalies, since these can be inferred from the context of the image even when no information is available for the specific patch we are reconstructing. For this reason, in future works, we would like to investigate more sofisticated masking approaches that allow for a better generalization of the model to different anomalies scales.
Finally in the future, we will also extend the our idea to process 3D applications \cite{Bergmann_2022}.

\section*{Acknowledgements}
This work was partially supported by the ONRG project N62909-20-1-2075 Target Re-Association for Autonomous Agents (TRAAA).
\bibliographystyle{ws-ijns}
\bibliography{main}


\end{multicols}
\end{document}